\documentclass{article}
\usepackage{spconf,amsmath,graphicx}

\usepackage{times}
\usepackage{epsfig}
\usepackage{graphicx}
\usepackage{amsmath}
\usepackage{amssymb}
\usepackage{amsmath,graphicx}
\usepackage{multirow} 
\usepackage{subfigure}
\usepackage{xcolor}
\usepackage{graphics}
\usepackage{epsf}
\usepackage{booktabs}
\usepackage{tabularx}
\usepackage{array,hhline}

\title{Learning zeroth class dictionary for human action recognition}
\name{Jia-xin Cai$^{1\ast}$ \qquad Xin Tang$^{2}$ \qquad Lifang Zhang$^{3}$ \qquad Guocan Feng$^{3\star}$ \thanks{This work is supported by Xiamen University of Technology High Level Talents Project (No.YKJ15018R) and the National Natural Science Foundation of P. R. China (No.61272338).}}

\address{$^{1}$ Xiamen University of Technology \ \  $^{2}$ Huazhong Agricultural University \ \  $^{3}$ Sun Yat-sen University \ \ \\
$^{\ast}$caijiaxin@xmut.edu.cn  \qquad  $^{\star}$ mcsfgc@mail.sysu.edu.cn}

\begin{document}
%
\maketitle
\begin{abstract}
 In this paper, a discriminative two-phase dictionary learning framework is proposed for classifying human action  by sparse shape representations, in which the first-phase dictionary is learned on the selected discriminative frames and the second-phase dictionary is built for recognition using reconstruction errors of the first-phase dictionary as input features.
 We  propose a "zeroth class" trick for detecting undiscriminating frames of the test video and eliminating them before voting on the action categories.
   Experimental results on benchmarks demonstrate  the effectiveness  of our method.
\end{abstract}
\section{Introduction}

Recently, human action recognition has gained much interest for its great  potential in many application areas such as video surveillance and human-computer interaction.
The challenge of human action recognition usually results from the problem that different action classes often share some common motion patterns.
Moreover, the action videos usually include many redundant frames indicating the background, large noise, clutter or small movements that are with limited help for recognition \cite{cai2013human}.
For action video classification, if the training and test videos contain some frames representing the common motion components among different action classes, or some redundant frames  with large noise or useless clutters, then the discriminability of  classifier learnt from training frames will be corrupted, and the recognition result of test video will also get corrupted when the labels of undiscriminating  test frames are used to determine the class label of test video.

Sparse coding based recognition approaches have attracted much attention in the field of computer vision \cite{Huang2014Feature}.
To learn a well-adapted dictionary for obtaining good reconstruction and recognition performance,
many algorithms \cite{Jiang2013,Yang2014,Shrivastava2014} for training dictionary with label information and discriminative criterion have been proposed.
These algorithms can only work well for the cleaning training samples or training samples with small corruption \cite{Li2014}.
So the efforts for learning low-rank and discriminative dictionary are made in \cite{Li2014,MaCVPR2012} to solve this problem.
However, the previous methods  can not handle the problem resulted from the undiscriminating test frame samples for video recognition.
Imagining the test video  contains many frames representing the useless clutters  or some common  components among different video classes,  the label of test video will get corrupted by these undiscriminating test frames.
So it is necessary to develop a discriminative dictionary for the situation that both the training and test videos contain many corrupted or uninteresting frame samples.

In this paper, we attempt to learn a discriminative dictionary for action recognition to handle the situation that both the training and test videos contain  undiscriminating  frames with common components, redundant components, background, clutter or large noise.
We  propose a recognition framework for human action recognition in video by learning a  discriminative dictionary called zeroth class dictionary.
 The "zeroth class" trick is proposed for detecting and filtering out the undiscriminating frames of the test video to eliminate the negative effect resulted from these frames during voting the action category of test video.
 The zeroth class dictionary method is a two-phase  dictionary learning system \cite{Bai2014Multiple,Li2011MULTI} including three steps:

 (1)Firstly, the discriminative frames of  training videos are selected  by Gentle Adaboost algorithm  to learn the first-phase  dictionary.
 The left undiscriminating frames are relabeled and assigned to the zeroth class.
The zeroth class is a virtual class indicating undiscriminating frames which are  with limited help for recognition,
such as frames with common poses shared by different actions, frames with clutter or noise, and other redundant frames.
Then the first-phase  dictionary is learnt on the  selected discriminative frames;
the reconstruction errors of all frames corresponding to each dictionary atom are collected to build the new frame representations.

(2)Using the new frame representations, we learn the class-specific dictionary in which the sub-dictionary of each action class is learned on the corresponding selected discriminative frames and the zeroth class sub-dictionary is learnt on the undiscriminating frames.
Then we obtain the preliminary labels of the test frames based on the learnt class-specific dictionary.
The zeroth class is used for recognizing the redundant frames of  test video, which are filtered out afterwards.

(3)After the undiscriminating frames of  test video are excluded, the final action label is voted by all remained discriminative frames of the test video.
Experimental results on benchmark data show that our method outperforms most state-of-the-art approaches.

The rest of the paper is organized as follows:
Sect.2 reviews the relative works.
Sect.3 presents the zeroth class dictionary learning framework
for human action recognition.
Experimental setup and results on
benchmark datasets are presented in Sect.4, and conclusions
are given in Sect.5.

\section{Relative works}
Many efforts \cite{Guha,ZhangICIP2015,Wang20123902} have been devoted to  studying  action recognition by sparse representation  and dictionary learning.
Guha and Ward \cite{Guha} provide a sparse representation for human action recognition by learning the over-complete bases on the local motion patterns.
Zhang et al.  \cite{ZhangICIP2015} learn dictionary from spatiotemporal salient patches and use the  sparse reconstruction coefficients  of patches to represent image sequences of action videos.
Wang et al. \cite{Wang20123902} propose a  sparse model incorporating the similarity constrained term and the dictionary incoherence term  for human action recognition.

Our work is also similar to the silhouette based action recognition approaches \cite{Chaaraoui2013,Cheema,frftacpr2015,ChengTIP2015}.
Chaaraoui et al. \cite{Chaaraoui2013} develop a human action recognition method through extracting multi-view key poses sequences and  handling variations in shape by dynamic time warping.
Cheema et al. \cite{Cheema}  propose a human action recognition method by extracting  a scale invariant contour-based pose feature  and  clustering  the features to construct distinctive key poses.
Cai and Feng \cite{frftacpr2015} present a human action recognition  method by describing contour-based shape feature using fractional Fourier transform.
Cheng  et al. \cite{ChengTIP2015} propose a human action recognition approach based on human silhouettes by supervised temporal t-stochastic neighbor embedding and incremental learning via low-dimensional embedding.

%
\section{Zeroth class dictionary learning based action recognition framework}

\subsection{Feature extraction}

We use the fractional Fourier shape descriptor \cite{frftacpr2015}  to represent each frame of action videos.
The fractional Fourier shape descriptor is built on the human pose represented by contour points of the binary silhouette.
Given an image extracted from the action video, its binary silhouette is obtained from the segmented foreground region.
Then the boundary of silhouette is extracted
 and the position of all points $\{(x(i),y(i))\}_{i=1}^{N}$ along the boundary is represented as a complex sequence $\{s(i)|s(i)=x(i)+jy(i)\}_{i=1}^{N}$, where $x(i)$ and $y(i)$ denote  the horizontal and vertical coordinate of the $i$th point respectively. Here $N$ is the total number of contour points, and $j$ denotes the imaginary unit.
Then we shift the base point of coordinate system to the center of mass $(x_{c},y_{c})$ of contour points along the boundary.
\begin{equation} \label{eq5}
 \begin{aligned}
\tilde{x}(i)=x(i)-x_{c} \ \ \ \ \ \  \tilde{y}(i)=y(i)-y_{c}
 \end{aligned}
\end{equation}

After that, the length of sequence is normalized to a predetermined value $L$ through down-sampling the contour.
In our experiments, the normalized length $L$ is set as 100.

 \begin{equation} \label{eq5}
 \begin{aligned}
\hat{x}(i)=\tilde{x}(\lceil i*\frac{N}{L} \rceil)  \ \ \ \ \ \  \hat{y}(i)=\tilde{y}(\lceil i*\frac{N}{L} \rceil)
\end{aligned}
\end{equation}

Afterwards, we compute the discrete fractional Fourier transform of the transformed contours $\{\hat{s}(i):\hat{x}(i)+j\hat{y}(i)\}_{i=1}^{L}$,
and get the response $\{S(i)\}_{i=1}^{L}$ in the fractional Fourier domain.
For a continuous signal $\hat{s}(t)$, its $p$ order continuous fractional Fourier transform is defined as:
\begin{equation} \label{eq2}
 \begin{aligned}
 S_{p}(u)=
  \begin{cases}
B_{\alpha}\int_{-\infty}^{\infty}exp(j\frac{t^{2}+u^{2}}{2}cot\alpha-\frac{jtu}{sin\alpha})\hat{s}(t)dt \\
\quad \quad \quad \quad \quad \quad \quad \quad \quad   \alpha \neq n\pi \\
\hat{s}(t)     \quad \quad \quad \quad \quad \quad \quad \alpha =2n\pi \\
\hat{s}(-t)    \quad \quad \quad \quad \quad \alpha =(2n\pm1)\pi
  \end{cases}
 \end{aligned}
\end{equation}
where $\alpha=p\pi/2$ is the rotation angle and  $B_{\alpha}=\sqrt{\frac{1-jcot\alpha}{2\pi}}$.
Here  $n$ denotes an integer.
The order $p$ is set as $0.9$ in our experiments.
For digital computation, we use a sampling-type discrete fractional
Fourier transform proposed in \cite{Ozaktas} to calculate  the response $\{S(i)\}_{i=1}^{L}$. 

Then the amplitude of fractional response, $|S(i)|$, is calculated and normalized to obtain a scale invariant descriptor.
 \begin{equation} \label{eq5}
 \begin{aligned}
d(i)=\frac{|S(i)|^{2}}{\sum_{i=1}^{L}|S(i)|^{2}}
\end{aligned}
\end{equation}

$\{d(i)\}_{i=1}^{L}$ consists the fractional Fourier shape descriptor of human pose contour.
For each training video, we assign its action class label to  its all affiliated frames.

\subsection{First-phase dictionary learning}
Firstly, the Gentle AdaBoost algorithm is  employed to select discriminative
training frames.
Gentle AdaBoost provides an  approach for reweighting data points by updating weights of base classifiers and  puts higher  weights on undiscriminating data points than  discriminative points \cite{Friedman98additivelogistic}.
Regression stump  is used as the base classifier. The regression stump is a simple additive logistic regression based classifier, which classifies data points according to only one input dimension. For an input sample $x$ whose $k$th dimensional feature is denote as $x(k)$, the output class label $f(x)$  of regression stump is defined by only four parameters $(w,v,k,th)$, and represented as follows.
\begin{equation} \label{eq5}
 \begin{aligned}
  f(x) = w*sign(x(k)-th)+v
 \end{aligned}
\end{equation}
The "one-against-the-rest" technique is employed to extend the primary binary
classification problem to multi-class case.
A training frame  with high weight imply that it contains common and undiscriminating patterns between different action categories.
We select the frames with the lowest weights from  the training frame set of each action class at a rate of $R$ to build the discriminative subset for generating the first-phase dictionary, and the remained frames are pushed into a pool where they are relabeled as the zeroth class and will be used for detecting
undiscriminating frames of the test video later.

After the discriminative subset of the training frames is selected out, we generate a dictionary $D$ on this set.
The aim is to learn a dictionary $D$ so that the selected discriminative frames  have a sparse representation $B$ over the dictionary.
It can be written as the following optimization problem  \cite{AharonKSVD}:
\begin{equation} \label{eq1}
 \begin{aligned}
min_{D,B}\|Y-DB\|_{F}^{2} \  \ \ \ \ \ \ \ \ s.t.\ \ |b_{i}\|_{0}\leq C ,\ \forall i
 \end{aligned}
\end{equation}

where $Y$ is the selected discriminative subset of training frames represented by the fractional Fourier descriptor;
$D$ is the learned dictionary on the discriminative subset; $b_{i}$ is the $i$th column of sparse coefficients matrix $B$, denoting the representing coefficient  of $i$th frame.
$C$ is the parameter controlling the sparsity of  coefficients.
$\|\cdot\|_{F}$ denotes the Frobenius norm, and $\|\cdot\|_{0}$  is the $l_{0}$ norm  enforcing the coefficients to be sparse.

Then for a frame $y$, the reconstruction error corresponding to the $i$th atom of the dictionary $D$ is computed as \cite{Pati93orthogonalmatching}:

\begin{equation} \label{eq1}
 \begin{aligned}
e_{i}(y)=\|y-D\delta_{i}(\hat{\beta})\|^{2}      \  \ \ \ \  \  \ \ \ \   \ \ \    \\
\hat{\beta}=argmin_{\beta}\|y-D\beta\|^{2}   \  \ \ \ \ s.t. \ \ \ \|\beta\|_{0}\leq C
 \end{aligned}
\end{equation}

where the function $\delta_{i}(\beta)$  sets the $j$th dimension of $\beta$ as 0 if $j\neq i$.
Suppose $m$ is the atom number of dictionary $D$,
then the vector $[e_{1}(y),...,e_{m}(y)]^{T}$ makes up a new feature of frame $y$, which would be used as the new frame feature in the next phase dictionary learning.

\subsection{Second-phase dictionary learning}
After the new features of all frames in both training and test videos are computed, the class-specific dictionary learning \cite{Wang20123902} is performed.
Suppose $K$ is the number of action categories.
Using the training frames belonging to the $k$th ($k$ $=$ $0,1,...,K$) class (including the zeroth class), we learn the class-specific dictionary $D_{k}$ using the new feature represented by reconstruction errors on the first-phase dictionary.
The sub-dictionary $D_{k}$ associated with the $k$th($k$ $=$ $1,...,K$) nonzero action class is learnt on the corresponding selected discriminative frames of the $k$th class, and the zeroth class dictionary $D_{0}$ is learnt on the undiscriminating frames.
Then the whole dictionary $\bar{D}$ is constructed by concatenating all the class-specific dictionaries, that is to say, $\bar{D}=[D_{0}|D_{1}| D_{2} |...| D_{K}]$.

After the whole dictionary $\bar{D}$ is learned, the sparse representation $a_{i}$ of a  frame  $\check{x}_{i}$ of the test video can be estimated as follows.
\begin{equation} \label{eq5}
 \begin{aligned}
a_{i}=argmin_{a}    \|\check{x}_{i}-\bar{D}a\|^{2}   \ \ \ \ \ \ \ \ \  s.t.  \ \   \|a\|_{0} \leq C
 \end{aligned}
\end{equation}
The reconstruction error $r_{k}(\check{x}_{i})$ associated with the $k$th class can be defined as:
 \begin{equation} \label{eq5}
 \begin{aligned}
r_{k}(\check{x}_{i})=\|\check{x}_{i}-\bar{D}\Theta_{k}(a_{i})\|^{2}   \ \ \ \ \ \ k=0,1,...,K
 \end{aligned}
\end{equation}
where $\Theta_{k}(a_{i})$ produces a vector whose nonzero entries are coefficients of $a_{i}$  associated with the $k$th class .

Then each frame of the test video is assigned to the class that corresponds to
the minimum of reconstruction error with respect to each class(including the zeroth class).
The estimated preliminary class  $\check{k}_i$ of the test frame $\check{x}_{i}$ is given as:
 \begin{equation} \label{eq5}
 \begin{aligned}
\check{k_i}=argmin_{k\in \{0,1,...,K\}} r_{k}(\check{x}_{i})
\end{aligned}
\end{equation}

 Afterwards, we  filtered out the undiscriminating frames in the test video which are labeled as the zeroth class.
Then the max pooling or sum pooling criteria is used  to vote the action label by the remained discriminative frames corresponding to nonzero classes of the test video.
For max pooling policy, each frame of the test video is classified to the nonzero class that corresponds to
the minimum of reconstruction error with respect to each non-zero class.
Then the estimated action class  $\hat{k}$ of the test video is given as:
 \begin{equation} \label{eq5}
 \begin{aligned}
\hat{k}=argmin_{k\in \{1,...,K\}}min_{\hat{i} \in \{i| \check{k_i} \neq 0  \}} r_{k}(\check{x}_{\hat{i}})    \\
\end{aligned}
\end{equation}

For sum pooling policy, an overall residual is constructed by summing up the reconstruction errors corresponding to nonzero classes of each frames in  the test video;
then the test video is assigned to the nonzero class with respective to the minimum of overall error.
The estimated action class  $\hat{k}$ of the test video is given as:
 \begin{equation} \label{eq5}
 \begin{aligned}
\hat{k}=argmin_{k\in \{1,...,K\}}\sum_{\hat{i} \in \{i| \check{k_i} \neq 0  \}} r_{k}(\check{x}_{\hat{i}})   \\
\end{aligned}
\end{equation}

\section{Experimental results}

In order to evaluate the performance and practicability of the proposed approach,
two human action recognition datasets, the Weizmann dataset \cite{ActionsAsSpaceTimeShapes_pami07} and the MuHAVi-MAS14 dataset \cite{Singh}, are used
as benchmarks.
For each class, we select frames with the lowest Gentle Adaboost weights as the discriminative subset at a rate of $R$.
The leave-one-out cross validation strategy is employed to separate the training video set and test video set.
All parameters are tuned by grid searching.
The best recognition rates on Weizmann dataset and MuHAVI-MAS14 dataset are achieved as 97.85\% and 95.59\% respectively when  $R$ is set as $0.2$ and  $C$ is set as $15$.
We also compare the accuracy of our method to the reported accuracy of  other state-of-the-art methods.
The comparison results are presented in Table1.
Although the action features employed in most listed methods are different to ours, our method still shows a considerable performance and outperforms most listed methods on the benchmarks.

\begin{table}[htbp]
  \centering
  \caption{Comparison of methods  on benchmarks}
    \begin{tabular}{p{4.5cm} p{1.5cm} p{1.5cm}}
    \toprule
    Method &  Weizmann &  MuHAVi-MAS14\\
    \midrule
    Our method(sum pooling) & 97.85\% & 95.59\%\\
   Our method(max pooling) & 95.70\% & 95.59\%\\
    Chaaraoui et al. \cite{Chaaraoui2013} & 92.77\% & 91.18\% \\
    Cheema et al. \cite{Cheema}  & 91.6\% & 86.03\%\\
        Singh et al. \cite{Singh}  &  & 82.35\%\\
       Wang et al. \cite{Wang20123902} & 96.7\% &  \\
    Cheng et al. \cite{ChengTIP2015} & 94.44\% &  \\
       Cai and Feng \cite{frftacpr2015} & 93.55\% &  \\
        \bottomrule
    \end{tabular}%
  \label{tab:recresult}%
\end{table}%

 Analysis of the relation between recognition accuracy and  the size of zeroth class set has also been carried out.
  Figure 1 and Figure 2 show the relation between accuracy and the rate $R$  on  Weizmann  and MuHAVI-MAS14 dataset respectively.
 Experiment results demonstrate that our framework outperforms the ordinary two-phase dictionary learning framework without the discriminative frame detection and filtering stage at most time.
However, if too many undiscriminating frames are selected into the zeroth class training set for the first-phase dictionary learning , the performance of the framework  will decline.
The experimental results  demonstrate introducing the zeroth class is  effective  if  a proper proration of zeroth class of the training set is set.
We have also analyzed  the relation between accuracy of our method and the parameter $C$.
The experimental results on Weizmann and  MuHAVI-MAS14 dataset are illustrated in Figure 3 and Figure 4 respectively.
The results demonstrate that it is easy to find a proper parameter $C$ for achieving good classification performance through the zeroth class dictionary learning framework.

\section{Conclusion}
This paper presents an action recognition method by using zeroth class dictionary.
The zeroth class dictionary provides a method to detect and delete undiscriminating
frames of test video for improving the classification accuracy.
The recognition framework is validated on benchmarks, showing a considerable performance.

\begin{figure}[htbp]
\centering
\includegraphics[height=3.8cm]{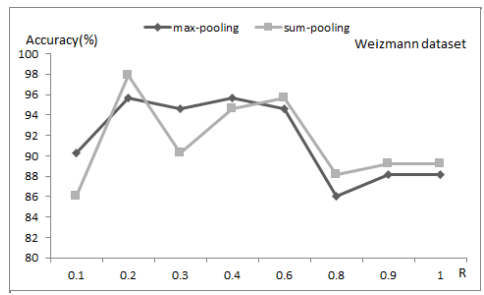}
\caption{Relation between accuracy  and the rate of selected discriminative frames on Weizmann dataset}
\label{Fig3}
\end{figure}

\begin{figure}[htbp]
\centering
\includegraphics[height=3.8cm]{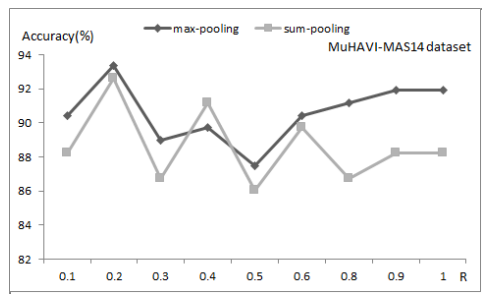}
\caption{Relation between accuracy  and the rate of selected discriminative frames on MuHAVI-MAS14 dataset }
\label{Fig3}
\end{figure}

\begin{figure}[htbp]
\centering
\includegraphics[height=3.8cm]{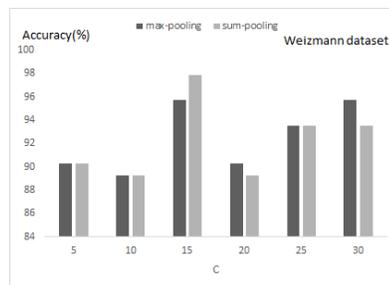}
\caption{Relation between accuracy  and parameter C on Weizmann dataset}
\label{Fig3}
\end{figure}

\begin{figure}[htbp]
\centering
\includegraphics[height=3.8cm]{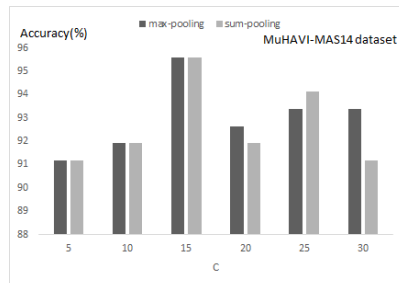}
\caption{Relation between accuracy and parameter C on MuHAVI-MAS14 dataset}
\label{Fig3}
\end{figure}

{
\bibliographystyle{IEEEbib}
\bibliography{egbib}
}

\end{document}